\documentclass[aps,pra,showpacs,preprintnumbers,amsmath,amssymb]{article}
\usepackage{arxiv}
\usepackage{mathptmx}
\usepackage{cite}  % *** CITATION PACKAGES ***
\usepackage[pdftex]{graphicx} % *** GRAPHICS RELATED PACKAGES ***
\usepackage{url}

\usepackage{times}
\normalfont

\title{
Individual specialization in multi-task environments with multiagent reinforcement learners
}

\author{	
	Marco Jerome Gasparrini\\
	University Pompeu Fabra\\
	Barcelona, Spain	 
	\And 
	Ricard Sol\'e\\
    ICREA-Complex Systems Lab\\
    University Pompeu Fabra\\
    Barcelona, Spain	 
	\And 
	Mart\'i S\'anchez-Fibla\thanks{Corresponding Author: Technology Department, Universitat Pompeu Fabra, Carrer de Roc Boronat 138, 08018 Barcelona, Spain. E-mail:
		marti.sanchez@upf.edu.}\\
	University Pompeu Fabra\\
	Barcelona, Spain	 
}

\renewcommand{\headeright}{arXiv ver.}
\renewcommand{\undertitle}{arXiv version\thanks{Version of the extended abstract presented in the International Conference of the Catalan Association for Artificial Intelligence, CCIA 2019 and appeared in 
		Artificial Intelligence Research and Development: Proceedings of the International Conference of the Catalan Association for Artificial Intelligence, Volume 319, pags 339-343, IOS Press, 2019.}}

\begin{document}
\maketitle

\keywords{Multiagents, Deep Reinforcement Learning, Multitask}

\section*{Extended Abstract}

There is a growing interest for Multi-Agent Reinforcement Learning (MARL) as first steps towards building general intelligent agents that learn to make low and high level decisions in non-stationary complex environments in the presence of other agents \cite{perolat2017multi,leibo2017multi,lerer2017maintaining, omidshafiei2017deep, freire2018modeling}. Previous results point us towards increased conditions for coordination, efficiency/fairness (see for example \cite{freire2018modeling} and  \cite{gasparrini2018loss,hughes2018inequity} in which loss averse agents are considered) and common-pool resource sharing \cite{perolat2017multi,hughes2018inequity}.
We further study coordination focusing on how individuals specialise in multitask environments where several rewarding tasks can be performed (as in \cite{omidshafiei2017deep}). Multitask environments are suited to study general intelligence \cite{legg2007universal}: the ability of agents to perform as well as possible in a big variety of tasks. 
Agents learning in big multi-agent populations don't necessarily need to perform well in all tasks, but under certain conditions may specialise. 
The number of individuals present in the learning may affect or be linked to the cognitive capabilities of each individual (see the relation of these two axis in \cite{arsiwalla2017morphospace}).
An example of this can be seen in large social insect communities in which individuals specialize and surprisingly can also switch tasks dynamically whenever needed as in ant colonies \cite{pinero2019statistical}.  

\begin{figure}[htbp]
  \begin{center}
   \includegraphics[width=6cm]{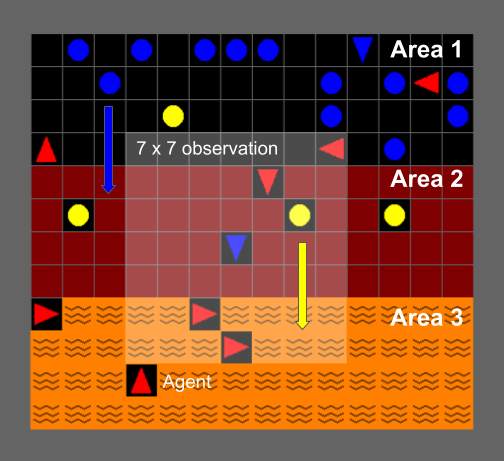}
   \caption{Grid-world Ant environment with agents being red triangles, resources in blue and yellow respectively belonging to the first and second tasks. Tasks consist in bringing blue resources to the red area two (Task 1) and yellow resources to the orange area three (Task 2).}
  \label{antenv}
   \end{center}
\end{figure}

We consider a multitask multi-agent environment in which  $n$ tasks can be performed sequentially, one after the other (we only use two tasks but could be generalized to any number). 
The agents have available resources in an area of the environment that need to be carried to the next area and when this is done the resource is assigned the next target area or disappears if in the last area. Concretely, agents will be rewarded for bringing resources of type one from the first area to the second (Task 1) and from the second to the third (Task 2), as shown in Figure \ref{antenv}. 
Agents can perform 5 actions (turn left and right, move forward, take and drop resource) and have partial observability of the environment (a square of 7x7 centered in its position). This partial observability contains several channels which determine what the agents can perceive: the ids of the different areas together with the walls, the types of resources, the other agent positions.
Agents get rewarded for the accomplishment of each task (when dropping a resource in the correct zone) with a reward value associated to the resource that decays over time.
The resources, like the agents, spawn at random positions at the beginning of each episode and their value resets when the first task is accomplished. We shaped the reward function in order to boost the learning speed, discouraging with negative rewards useless actions or waiting behaviours.

\begin{figure}[htbp]
  \begin{center}
   \includegraphics[width=6.1cm]{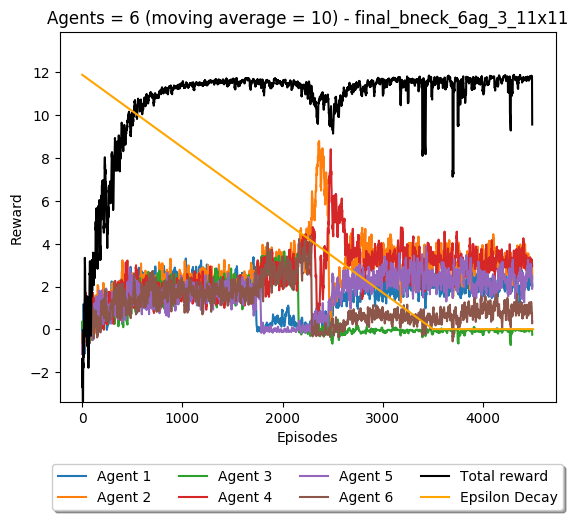}
    \hspace{.01cm}
    \includegraphics[width=6.1cm]{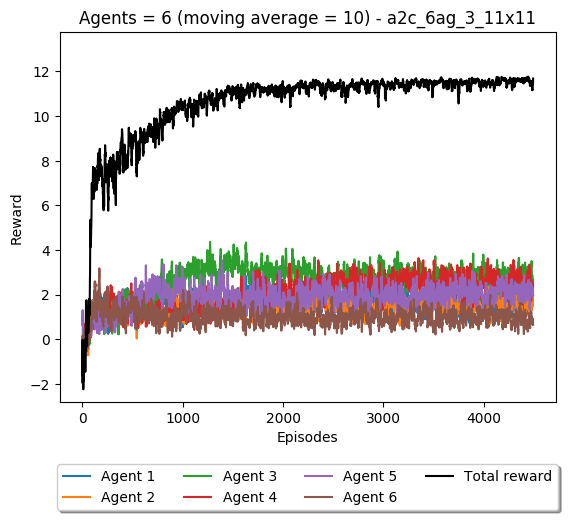}
  \end{center}
  \caption{Ant Environment. \textbf{Left:} Example of one session of the simulation with 6 agents using DDDQN. Reward of each agent and total reward (in black) is plotted per episode. The linear decaying orange line indicates the amount of random actions the agents perform during learning ($\epsilon$-decay). As we point out in the text, the global synchronization among agents caused by the shared $\epsilon$ probability (epsilon decay) of taking a random action forces all agents to act deterministically at the same time and creates instabilities and difficulties to converge. \textbf{Right:} Same simulation using A2C. No $\epsilon$-decay is present in policy gradient based algorithms.}
  \label{sim}
\end{figure}

We study factors that affect specialisation such as agent number, task throughput (limiting the number of times one task can be performed before another task: in our case this means limiting the number of resources that can be in area two, we also call it bottleneck) and reward degradation over time. Agents are independent learners, trained through two state of the art reinforcement learning algorithms: Dueling Double Deep Q-Network (DDDQN) \cite{van2016deep,DBLP:journals/corr/WangFL15} with agent-independent prioritised experience replay buffers \cite{schaul2015prioritized} and Asynchronous Advantage Actor-Critic (A2C) \cite{mnih2016asynchronous}, both using Convolutional Neural Networks as function approximators. The former is a value-based method, that learns a value function and then acts greedily with respect to it. The policy is intrinsically given by the value function itself and the actions are selected according to a time-decreasing probability called epsilon ($\epsilon$-greedy) that controls whenever they will be sampled according to the current estimated value function or from a uniform random distribution ($\epsilon$ can be seen to decrease linearly and equally for all agents in Figure \ref{sim}). On the other hand, the algorithm A2C, is a mix of value based and policy gradient approaches, in which we have two networks, an actor who maps states to actions and a critic (value based) who learns how well the actor is performing and contributes to improve the policy. This is performed through the actor policy gradient, who aims to change the policy to maximize the value estimations of the critic. In this case the exploration is controlled by the entropy of the policy (how uncertain and similar to a uniform distribution it is), computed in the actor network loss function, without any external parameters.

\begin{figure}[htbp]
  \begin{center}
    \includegraphics[width=6.1cm]{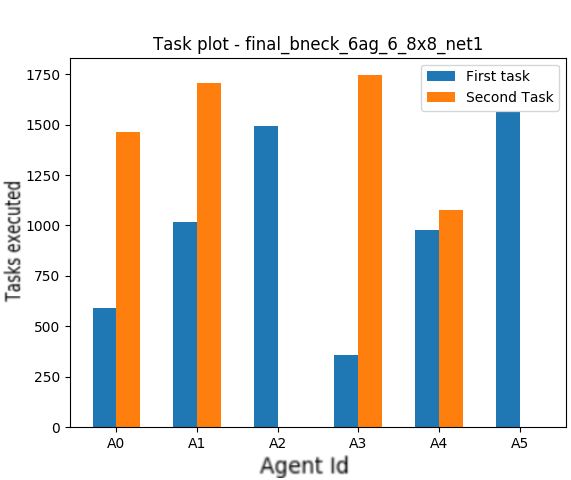}
    \hspace{0.00001cm}
    \includegraphics[width=6.1cm]{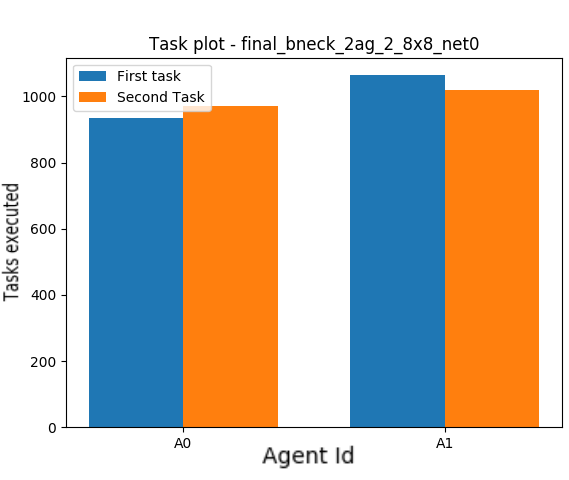}
  \end{center}
  \caption{Agent specialization outcomes. Simulations on 8x8 grids. Bars represent number of Task1 and Task2 being executed over 1000 episodes. \textbf{Left:} Two agents learn to perform both tasks in a very fair and shared way. \textbf{Right:} Outcome of the specialization for a simulation with six agents in an 8x8 grid.  The division of labor per task is much more heterogeneous and more specialized.}
  \label{spec}
\end{figure}

The specialization $S_a$ of each agent $a$ is defined as the absolute difference of the two tasks counts (number of times a task has been done) $t_1$ and $t_2$ divided by their sum: $S_a = \frac{|t_1 - t_2|}{t_1+t_2}$. In this way we will consider an agent more specialized when it mainly performs one task independently of its respawn location. 

We expect to observe a progressive specialisation in single tasks with respect to agent number. This result would indicate the advantage of sacrificing general cognitive abilities for the benefit of population efficiency as it happens in ant colonies \cite{pinero2019statistical}.

The results we obtained using independent DDDQN learners suggest that, increasing the number of agents, increases their specialization (see Figure \ref{spec}).
Nevertheless, DDDQN is unstable and has difficulties to converge (we measured convergence by looking at the weight updates of the network).
On the other hand, using A2C proved to be much more stable and all agent networks converged. We attribute the enhanced convergence of A2C with respect to DDDQN to the fact that the two algorithms have very different exploration strategies. In the case of DDDQN, all agents are synchronized by the external exploration factor $\epsilon$ which is the probability of taking a random action. This causes an undesirable situation as all agents are forced to have deterministic policies at the same time and they cannot adapt to each other, thus creating the instabilities and the convergence problems. This is not the case of A2C in which exploration is determined by the action probability distributions of the actor network and vary across agents independently. 

Another problem of DDDQN may be its experience replay. In fact it can lead to more non-stationarity since experiences related to the other agents' behaviour collected in the past and stored in the buffers might be replayed to the network when they are outdated \cite{omidshafiei2017deep}. We can have an idea of this effect looking at Figure \ref{sim}; the two different plots show respectively the same simulation performed using the two DDDQN and A2C. The former, that uses $\epsilon$-greedy as exploration strategy makes all the agents explore at the same time, leading to a drop of the total community reward due to competition when the policy starts to be more deterministic. The total reward will be recovered only when a new equilibrium will be reestablished among the agents. This reason is also probably responsible of the further instability, that is not noticeable in the A2C simulation plot, in which the exploration is connected to the entropy level of the policy, allowing the agents to explore not necessary all at the same time with the same rate. In addition, the positive correlation between number of agents and specialization is confirmed and clearer in the experiments performed with A2C.     

The throughput (also called bottleneck) seems to affect fairness and the stability of the learning, probably because the environment will be more difficult to exploit. This may be another reason why the simulations with a stricter (lower) bottleneck had more problems to converge.

The next steps will be to employ A2C algorithm exclusively and perform a wider set of simulations with different environment sizes, number of agents and resources availability. A2C has no convergence problems and we expect to obtain a cleaner specialization trend. Then we could probably be able to link the throughput and the agent number with the specialization and study if there will be a positive correlation, as we expect. 

\vspace{.3cm}
\noindent \textbf{Acknowledgments}
\begin{small}
Research supported by INSOCO-DPI2016-80116-P. 
\end{small}  

\vspace{-.3cm}

\bibliographystyle{plain}
\bibliography{specialization}

\end{document}